# Evaluation Before Generation: A Paradigm for Robust Multimodal Sentiment Analysis with Missing Modalities

Rongfei Chen[1], Tingting Zhang[2], Xiaoyu Shen[1], Wei Zhang[1*]
[1]School of Computer Science and Technology, Eastern Institute of Technology, Ningbo, China
[2]School of Mechatronic Engineering and Automation, Shanghai University, China
{rfchen, xyshen, zhw}@eitech.edu.cn, tt302@shu.edu.cn

*Abstract*—The missing-modality problem constitutes a fundamental challenge in multimodal sentiment analysis, critically compromising model accuracy and generalization stability in real-world applications. Current research predominantly enhances model robustness against missing modalities through independent prompt learning and pre-trained models. However, two key limitations persist: (1) the necessity of generating missing modalities lacks rigorous evaluation; (2) the structural dependencies among multimodal prompts and their global coherence remain insufficiently studied. To address these limitations, we propose a Prompt-based Missing Modality Adaptation framework (ProMMA). Specifically, a Missing Modality Evaluator (MME) is embedded at the data input stage to dynamically assess the importance of missing modalities via pretrained models and constructed pseudo-labels, thereby avoiding the generation of low-quality imputed data. Based on this, a Modality-invariant Prompt Disentanglement (MiPD) module disentangles shared prompts into modality-specific private prompts, enhancing data generation quality by capturing intrinsic local prompt correlations. Additionally, the Dynamic Prompt Weighting (DPW) module computes mutual information weights from cross-attention outputs, mitigating interference from missing modalities via adaptive prompt weighting. To propagate global prompt semantics, a Multi-level Prompt Dynamic Connection (MPDC) module integrates shared prompts with self-attention outputs via residual connections, leveraging the global prior of shared prompts to reinforce critical guidance features. Extensive experiments on three public MSA benchmarks (CMU-MOSI, CMU-MOSEI, and CH-SIMS) demonstrate that ProMMA maintains state-of-the-art accuracy and stability across diverse missing-modality scenarios. The code implementation of this work is publicly available on GitHub: https://github.com/rongfei-chen/ProMMA.

*Index Terms*—Missing-Modality Adaptation, Prompt Learning, Multimodal Sentiment Analysis

## I. Introduction

Multimodal sentiment analysis (MSA) integrates data from text, audio, and visual modalities to enable models to recognize and respond to human emotions. As a result, it is widely used in areas such as embodied interaction, mental health support, and recommendation systems [1]. Currently, most multimodal sentiment analysis methods assume the availability of all modalities, treating them as complementary sources of information. However, this assumption overlooks the issue of

This work is supported by 2035 Key Research and Development Program of Ningbo City under Grant No.2024Z127.
*Corresponding author: Wei Zhang (zhw@eitech.edu.cn).

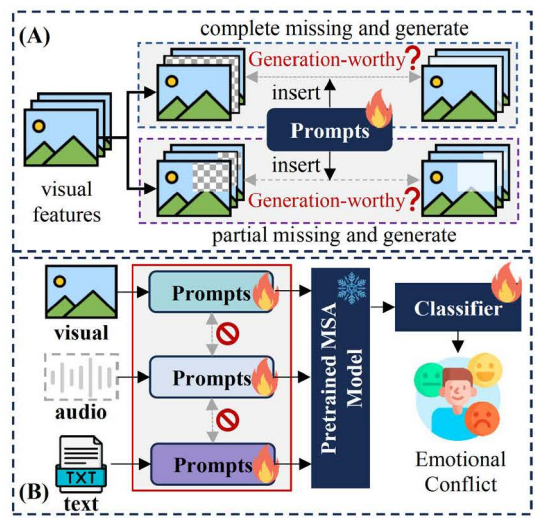

Fig. 1. Problem Formulation: (A) Patterns of incomplete multimodal data and Problem 1: whether generating missing data is justified. (B) Problem 2: information barriers between modality-specific prompts.

missing or low-quality modalities due to real-world factors such as human error, privacy constraints, sensor malfunctions, or environmental changes, which can significantly degrade model performance [2].

To address the issue of missing modalities, existing research focuses on three main approaches: data generation, shared space learning, and knowledge transfer [3]. Data generation methods improve multimodal data completeness by synthesizing missing modal information, thereby enhancing cross-modal interaction reliability [4]. In contrast, shared feature learning methods enforce multimodal data consistency, guiding the model to learn shared representations across modalities in latent space and reducing uncertainty caused by missing data [5]. Knowledge transfer techniques, including knowledge distillation, leverage task-relevant prior knowledge from complete multimodal data and employ transfer or distillation strategies to mitigate the effects of missing modalities [6]. However, these methods primarily focus on generating valid missing data without considering the importance of the missing modality for downstream tasks, particularly whether the missing modality is "worth completing" (shown in Fig.1-A). Additionally, existing prompt learning methods primarily construct independent, modality-specific or task-specific prompts, overlooking the consistency of modal soft prompts within the "emotional semantic space" (shown in Fig.1-B).

This study addresses the issue of missing modalities in multimodal sentiment analysis through prompt learning, focusing on two key questions: 1) Is it valuable to generate the missing modality? 2) How can consistency priors be used to improve prompt quality and enhance generation accuracy? To tackle these challenges, we propose a novel framework for missing modality adaptation based on prompt learning, named ProMMA. We introduce a missing modality evaluator to assess the "missing quality" of multimodal inputs before they are processed by a pre-trained backbone network, enabling the model to determine whether generating the missing modality is necessary. Additionally, we design three modules: a Modality-invariant Prompt Disentanglement (MiPD) module, a Dynamic Prompt Weighting (DPW) module, and a Multi-level Prompt Dynamic Connection (MPDC) module. These modules investigate the relationships between different prompts by focusing on prompt decoupling, dynamic weighting, and multi-level fusion, respectively, thus enhancing the model's robustness to missing modalities.

## II. RELATED

Multimodal Learning with Missing Modalities (MLMM), or Incomplete Multimodal Learning (IML), confronts two primary challenges: modality heterogeneity and robustness to missing data. Recently, prompt learning, particularly soft prompt learning, has been increasingly applied to multimodal tasks to enhance the downstream performance of pre-trained models [7]. According to the structure of the prompt, these approaches can be categorized into hard prompt-based and soft prompt-based methods in MLMM. Hard prompts [8] are readable text strings meticulously crafted by human experts using natural language vocabulary, with the objective of guiding pre-trained models to acquire specific knowledge required for downstream tasks. This method is structurally simple, highly interpretable, and training-free. But its design is heavily reliant on expert knowledge, rendering hard prompts highly sensitive to the specific downstream task. Soft prompts [9] consist of a set of learnable continuous high-dimensional vectors. These vectors are embedded into the model input to participate in the fine-tuning and optimization of the pre-trained model. Leveraging the learnable nature of soft prompts, this approach can better adapt to task objectives, thereby potentially enhancing model efficiency and performance on downstream tasks. However, the soft prompt method may encounter prompt conflict issues in addressing missing modality, which can lead to increased model complexity and performance degradation in certain scenarios. Built upon a soft prompt foundation, the ProMMA framework utilizes prompt decoupling and dynamic connections to mitigate potential prompt conflicts. Moreover, a concurrent prompt weighting strategy reinforces the soft prompt's efficacy and adaptability, thereby enhancing the framework's robustness to missing modalities.

## III. METHODOLOGY

This section elaborates on the proposed Prompt-based Missing Modality Adaptation (ProMMA) framework for evaluating and tackling missing modality in MSA. As shown in Fig.2, the proposed framework comprises four key components: a Missing Modality Evaluator (MME), a Modality-invariant Prompt Disentanglement module (MiPD), a Dynamic Prompt Weighting module (DPW), and a Multi-level Prompt Dynamic Connection module (MPDC).

### A. Overall Framework

The proposed ProMMA framework adopts the pre-trained MULT [10] as its backbone network $\mathcal{M}_{base}$. We denote the multimodal dataset as $\mathcal{D} = (\mathcal{X}_1, \mathcal{X}_2, ..., \mathcal{X}_n)$, where each sample $\mathcal{X}_i$ comprises three modal representations: $\mathcal{X}_i = (\mathcal{X}_i^a, \mathcal{X}_i^v, \mathcal{X}_i^t)$. To formalize the representation of missing modalities, the symbol $\bar{*}, * \in \{a, v, t\}$ is used to indicate a missing modality. For instance, $\bar{a}$ denotes missing audio.

Within this framework, we incorporate an evaluator $\mathcal{G}_{eval}$ alongside multiple prompt optimization modules $\mathcal{G}_*^{Pro}, * \in \{a, v, t\}$ to strengthen the model's robustness in handling missing modalities. Specifically, multimodal missing data $\mathcal{X}_i^{\bar{a}, v, t}$ is first assessed by a evaluator to determine its necessity for generation. If generation is required, the missing modality is reconstructed via a modality-invariant prompt decoupler $\mathcal{M}_{\bar{a}, v, t}^{dec}$; otherwise, the original incomplete data are fed directly into the backbone network.

Similar to MPLMM [11], the $\mathcal{M}_{\bar{a}, v, t}^{dec}$ comprises a modality-specific prompt generator $\mathcal{G}_{\bar{a}}^{Pro}$ and multiple cross-modal auxiliary generators $\mathcal{G}_{t/v \rightarrow a}^{cross}$. It differs in generating prompts for different modalities through complete prompts $\mathcal{P}_{\bar{a}, v, t}^{COM}$ decoupling, thereby enhancing potential global contextual relationships among modality-specific prompts.

To mitigate information loss and disturbances caused by missing modalities, the dynamic prompt weighting module $\mathcal{M}_{\bar{a}, v, t}^{DPro}$ adaptively generates modality-specific matrix weights $\sigma_{a, v, t}$ based on cross-modal mutual information, reweighting the weight prompts $\mathcal{P}_{\bar{a}, v, t}^{WEI}$ to produce weighted prompts. Meanwhile, the multi-level prompt dynamic connection module $\mathcal{M}_{\bar{a}, v, t}^{CPro}$ optimizes missing prompts through weighted processing and concatenates the $\mathcal{P}_{\bar{a}, v, t}^{COM}$ via residual connections. This approach preserves the prior semantics of global prompts while preventing the loss of critical guidance information. The specific descriptions of each module are as follows.

### B. Missing Modality Evaluator

The missing modality evaluator $\mathcal{G}_{eval}$ aims to quantify the impact of generating missing modalities on the performance of downstream tasks. It comprises a pre-trained model $\mathcal{M}_{base}$ and a trainable multi-layer MLP classifier $\mathcal{G}_{mlp}$. The training process consists of two stages: first, the pretrained model $\mathcal{M}_{base}$ serves as a classifier, taking both complete and corresponding missing modality data as input to generate sentiment predictions, ie., $\hat{y}_{pred} = \mathcal{M}_{base}(\mathcal{X}_n^{a, v, t})$ and $\hat{y}_{miss} =$ $\mathcal{M}_{base}(\mathcal{X}_n^{\bar{a},v,t})$. The evaluator then computes the similarity $f_{sim}$ between these two prediction results and compares it with a manually defined threshold $\epsilon = 0.3$ to construct pseudo-labels $\tilde{y}_{pse}$ representing the degree of modality missingness. This process is formalized as follows:

$$\tilde{y}_{pse} = max(0, f_{sim}(\hat{y}_{pred}, \hat{y}_{miss}; \epsilon)) \quad (1)$$

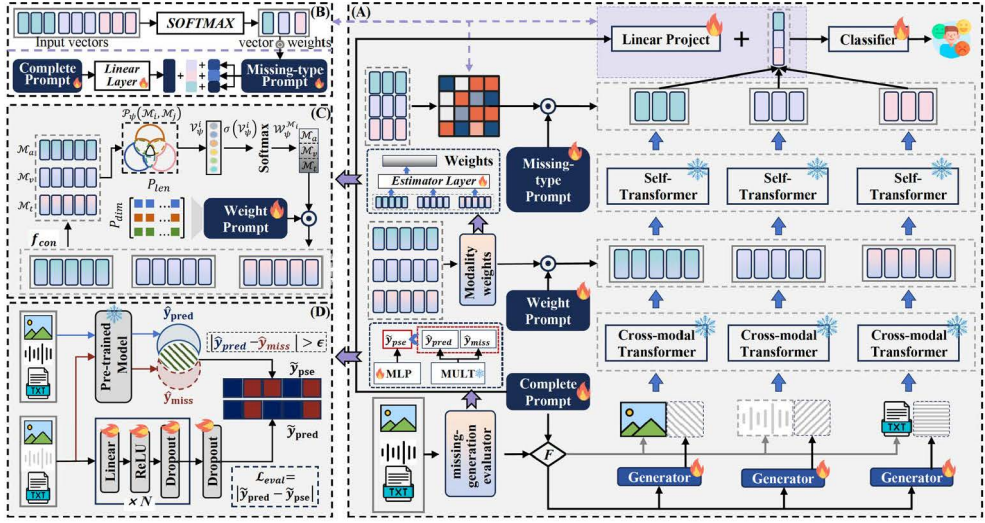

Fig. 2. Framework overview. (A) Pretrained backbone and core modules of the ProMMA; (B) schematic of the Multi-level Prompt Dynamic Connection (MPDC) module; (C) schematic of the Dynamic Prompt Weighting (DPW) module; (D) schematic of the Missing Modality Evaluator (MME) module. Here, $F$ denotes the missing-generation threshold.

Second, the same input $\mathcal{X}_n^{\bar{a},v,t}$ is used as input to train the classifier $\mathcal{G}_{mlp}$ to predict these pseudo-labels $\tilde{y}_{pse}$. Upon completion of training, the optimal model parameters are frozen and integrated into the backbone network.

### C. Modality-invariant Prompt Disentanglement

To generate missing modalities, we adopt the similar data generation structure as MPLMM. Specifically, MiPD employs multiple cross-modal MLP transformation layer to project features from available modalities into the feature space of the missing modality. The inter-modal data generation and transformation can be formulated as:

$$\mathcal{X}_i^{\bar{a}} = \mathcal{F}_a^{t/v}\left(X_i^{t/v}\right) = GELU(Conv1D_{d_{t/v} \to d_a}(X_i^{t/v})) \quad (2)$$

where $X_i^{t/v} \in \mathbb{R}^{d_l \times d_f}$, $d_l$ and $d_f$ denote the feature dimension and sequence length of the available modality, respectively. The Conv1D layer maps inputs from available complementary modalities to the target missing modality, achieving cross-modal feature dimension alignment. The GELU activation function is applied to introduce non-linearity, producing the transformed feature sequence. To enhance task consistency and intrinsic cross-modal prompt connections, $\mathcal{G}_*^{Pro}, * \in \{a, v, t\}$ generates modality-specific prompts $\mathcal{P}_*^{Spec}$ by decoupling shared complete prompts $\mathcal{P}_{\bar{a},v,t}^{COM}$:

$$\mathcal{P}_*^{Spec} = \mathcal{G}_*^{Pro}(\mathcal{P}_*^{COM}), * \in \{a, v, t\} \quad (3)$$

Structurally, the decoupling generator comprises multiple MLP layers, with identical network architectures used across different modalities. Overall, the MiPD output includes decoupled prompts and generated features from complementary modalities. Furthermore, a dimension mapping module $\mathcal{F}_\phi$ based on multi-layer MLPs is constructed to align the feature dimensions of different modality features and prompts, ensuring consistent dimensionality across heterogeneous data. The output of MiPD can be defined as:

$$\mathcal{X}_{\bar{a}}^{'} = \mathcal{F}_\phi([\mathcal{P}_a^{Spec}, \mathcal{X}_i^{\bar{a}}]), \mathcal{X}_v^{'} = \mathcal{F}_\phi(\mathcal{X}_i^v), \mathcal{X}_t^{'} = \mathcal{F}_\phi(\mathcal{X}_i^t) \quad (4)$$

### D. Dynamic Prompt Weighting module

Given the information disparity between missing and complete modalities, the DPW module $\mathcal{M}_{\bar{a},v,t}^{DPro}$ computes the mutual information among the cross-modal Transformer output vectors, which is then used as weighting information for the prompts $\mathcal{P}_{\bar{a},v,t}^{WEI}$. Specifically, InfoNCE is employed to estimate the lower bound of mutual information, formulated as:

$$\mathcal{I}(\mathcal{X}_i^{'}; \mathcal{X}_j^{'}) \geq LogK - \mathcal{L}_{infoNCE}, i \neq j; i, j \in \{a, v, t\} \quad (5)$$

where K=*batch_size-1* is the number of negative samples, and $\mathcal{L}_{infoNCE}$ represents the InfoNCE loss computed with intra-batch negatives, cosine similarity, and temperature $\tau = 0.1$.

Drawing on the symmetry property of mutual information, the bimodal mutual information weight is defined as the mean of the mutual information between the two modalities. For instance, the visual-acoustic mutual information weight $\mathcal{I}_{a,v}$ is obtained by averaging $\mathcal{I}(\mathcal{X}_a^{'}; \mathcal{X}_v^{'})$ and $\mathcal{I}(\mathcal{X}_v^{'}; \mathcal{X}_a^{'})$, i.e., $\mathcal{I}_{a,v} = mean(\mathcal{I}(\mathcal{X}_a^{'}; \mathcal{X}_v^{'}), \mathcal{I}(\mathcal{X}_v^{'}; \mathcal{X}_a^{'}))$. To achieve dynamic weight allocation, softmax normalization is applied to convert the raw mutual information values into a comparable set of probability distributions. Subsequently, the weighted prompt $\mathcal{P}_*^{DWEI}, * \in \{a, v, t\}$ is derived by performing element-wise multiplication between each modality's mutual information weight and the shared weight prompt matrix. Taking missing modality $\mathcal{X}'_{\bar{a}}$ and available modalities $\mathcal{X}'_t$ as examples, the weighted prompt calculation proceeds as follows:

$$\mathcal{P}^{DWEI}_{\bar{a}} = \mathcal{I}_{a,t} \odot \mathcal{P}^{WEI}_{\bar{a},v,t} \qquad (6)$$

Based on the weighted prompts, the input to the self-attention Transformer for missing acoustic modality can be expressed as $\mathcal{X}''_{\bar{a},v,t} = \mathcal{P}^{DWEI}_{\bar{a}} \oplus \mathcal{X}'_{\bar{a}}$.

### E. Multi-level Prompt Dynamic Connection module

The core of $\mathcal{M}^{CPro}_{\bar{a},v,t}$ involves dynamically weighting missing-type prompts derived from feature vectors and establishing global connections based on complete prompts $\mathcal{P}^{COM}_*, * \in \{a, v, t\}$. Similar to the DWP module, the MPDC applies a softmax layer to probabilistically normalize the self-attention Transformer outputs and utilizes the identical weighting method to generate weighted missing-type prompts for each modality. Notably, normalization along the feature channel dimension for adaptive multimodal fusion weights:

$$\mathcal{P}^{wmt}_{\bar{a}} = softmax(\mathcal{X}''_{\bar{a}}; \sum(\mathcal{X}''_{\bar{a}/v/t})) \qquad (7)$$

Subsequently, vector expansion, element-wise matrix addition, and vector concatenation are applied to obtain the optimized weighted prompt output vector:

$$\mathcal{O}_* = concat(\mathcal{P}^{ext}_* \oplus \mathcal{X}''_*), * \in \{a, v, t\} \qquad (8)$$

To preserve the prior semantics of the global prompt while capturing key prompt information across different stages, the MPDC module employs a residual connection $f_{rc}$ to concatenate the complete prompt $\mathcal{P}^{COM}_*, * \in \{a, v, t\}$ with the output vector $\mathcal{O}_*, * \in \{a, v, t\}$. Notably, a trainable linear mapping layer $\mathcal{F}_\psi$ with parameter $\theta$ is incorporated to align the prompt dimensions, ensuring vector consistency. This final outpu is defined as:

$$\mathcal{O}^{cls}_* = f_{rc}(\mathcal{F}_\psi(\mathcal{P}^{COM}_*; \theta), \mathcal{O}_*), * \in \{a, v, t\} \qquad (9)$$

It is important to emphasize that, to maintain compatibility across the different modules, no modifications were made to the loss functions or parameter update procedures of the backbone framework. Ultimately, the $\mathcal{O}^{cls}_{a,v,t}$ are fed into the classifier to yield the final prediction.

## IV. EXPERIMENT

### A. Experimental setup

*1) Benchmark Data.:* Our experiments are conducted on three established MSA datasets: CMU-MOSI [12], CMU-MOSEI [13], and CH-SIMS [14]. Notably, the backbone network and missing modality estimator are pre-trained on each benchmark dataset. Subsequent evaluation of ProMMA is conducted using these pre-trained networks.

*2) Evaluation Metrics.:* In this study, binary classification accuracy (ACC) and F1 score (F1) are used as comprehensive performance evaluation metrics. In addition, we use MAE and CORR metrics to evaluate the model performance under different parameter (or module) conditions.

*3) Baselines.:* To evaluate the effectiveness and robustness of DPMSA, we compared it with seven representative baseline methods. Specifically, we introduced three different missing data handling strategies: lower bound (LB) [11], missing value substitution (MS) [11], and modality dropout (MD) [11], which reduce the model's sensitivity to missing data from the perspectives of independent training, data substitution, and data removal, respectively. In addition, we introduced four SOTA methods: MCTN [15], MMIN [16], MPMM [17], and MPLMM [11], which address the problem of modality missing in multi-scenario settings using joint representation learning and prompt learning, respectively.

### B. Main Results

Based on three benchmark datasets, we evaluated ProMMA under six missing-modality scenarios to comprehensively assess model performance. Our method only trains lightweight soft prompts (length 39/50/50) and 30 MLP/Conv1d-based layers, resulting in a total of 1.30M trainable parameters, compared with 0.70M parameters in the pre-trained backbone. The extra overhead is less than 45.9% of the full model (-1.24s per batch sample inference) while our method achieves 1.5% performance gain in 30% missing-modal scenarios. Table I illustrates that ProMMA achieves the best overall performance across diverse benchmark datasets and missing-data scenarios, underscoring its efficiency and robustness in addressing missing-modality problems. Specifically, ProMMA achieves the best average performance across all three benchmark datasets: MOSI (ACC=78.04%, F1=77.50%), MOSEI (ACC=81.75%, F1=81.32%), and CH-SIMS (ACC=74.89%, F1=77.74%).

Notably, MPLMM marginally outperforms the ProMMA model on certain metrics: F1{a,t}=81.09% on MOSI, and ACC{v}=79.75% / F1{v}=78.86% on CH-SIMS. This difference may be attributable to modality imbalance and variations in prompt information. Specifically, the missing-signal prompt in MPLMM guides the model to identify input data attributes (i.e., generated or original), while its prompt matrix method supplies rich multimodal semantic information, making it potentially more suitable for dominant-modality scenarios. However, both the generated data and the prompt matrix may introduce substantial redundancy, which could elevate uncertainty during multimodal fusion and thus lead to suboptimal performance across most metrics. From a methodological perspective, the MME in ProMMA functions analogously to the missing-signal prompt, guiding the model to accurately identify missing data attributes. Furthermore, the MiPD reduces the informational uncertainty in prompt matrices, enabling the model to prioritize the generation quality of missing modalities.

### C. Ablation Results and Analysis

*1) Contributions of different modules.:* The ablation results in Table II demonstrate consistent performance gains as each module is successively integrated into the backbone architecture. Specifically, the MME and MiPD modules jointly enable the model to evaluate the significance of missing modalities and filter interference from low-quality data. The prompt architecture, which integrates globally shared prompts with modality-specific private prompts, provides comprehensive guidance while remaining adaptable to the characteristics of each individual modality. Incorporating the DPW module yields a further performance improvement. This indicates that cross-attention effectively captures and balances inter-modal information, while DPW refines the underlying distribution of modalities to calibrate their feature contributions. The integration of the MPDC module allows the model to utilize global prompt priors and integrate coarse-grained macro-features within the multimodal fusion process, thereby achieving further performance improvements. Together, these modules constitute a robust and adaptive framework.

TABLE I
MODEL PERFORMANCE UNDER SIX MISSING MODALITY SCENARIOS ON THE BENCHMARK DATASET. NOTE: ∗, ∗ ∈ {a, v, t} DENOTES AVAILABLE MODALITIES; "AVG." INDICATES AVERAGE PERFORMANCE ACROSS MISSING-MODALITY CONDITIONS; **BOLD** VALUES SIGNIFY BEST RESULTS. THE MISSING RATE AND EVALUATION THRESHOLD WERE FIXED AT 30%. ALL COMPARATOR RESULTS ARE SOURCED FROM [11]

| Dataset | Method | {a} | | {v} | | {t} | | {a,v} | | {a,t} | | {v,t} | | avg. | |
|---|---|---|---|---|---|---|---|---|---|---|---|---|---|---|---|
| | | ACC | F1 | ACC | F1 | ACC | F1 | ACC | F1 | ACC | F1 | ACC | F1 | ACC | F1 |
| MOSI | LB | 48.32 | 55.81 | 49.09 | 55.20 | 79.27 | 79.22 | 50.07 | 57.12 | 57.12 | 79.25 | 79.86 | 79.96 | 64.21 | 67.76 |
| | MS | 49.17 | 55.34 | 49.87 | 56.12 | 78.06 | 78.28 | 51.12 | 57.01 | 79.32 | 79.65 | 80.32 | 80.38 | 64.64 | 67.80 |
| | MD | 48.79 | 55.74 | 49.66 | 55.60 | 79.36 | 80.01 | 52.33 | 56.84 | 79.59 | 79.86 | 80.51 | 80.43 | 65.04 | 68.08 |
| | MCTN | 51.32 | 56.12 | 54.27 | 56.33 | 79.63 | 79.78 | 56.79 | 57.84 | 78.96 | 79.17 | 80.45 | 80.65 | 66.90 | 68.32 |
| | MMIN | 59.16 | 60.12 | 61.01 | 61.98 | 80.10 | 80.16 | 63.79 | 64.08 | 80.50 | 80.33 | 80.46 | 80.63 | 70.84 | 71.22 |
| | MPMM | 57.26 | 59.35 | 58.63 | 59.12 | 79.81 | 80.10 | 60.54 | 61.33 | 79.89 | 79.84 | 80.74 | 80.93 | 69.48 | 70.11 |
| | MPLMM | 62.71 | 63.65 | 63.12 | 63.74 | 80.12 | 80.31 | 65.02 | 65.41 | 80.76 | **81.09** | 81.12 | 81.19 | 72.14 | 72.57 |
| | ProMMA | **74.54** | **73.63** | **75.93** | **74.84** | **80.32** | **80.65** | **75.00** | **73.41** | **80.93** | 80.84 | **81.54** | **81.63** | **78.04** | **77.50** |
| MOSEI | LB | 66.21 | 68.69 | 66.45 | 69.10 | 77.96 | 78.32 | 67.30 | 69.62 | 78.13 | 78.63 | 77.86 | 78.16 | 72.32 | 73.83 |
| | MS | 62.74 | 67.06 | 64.16 | 68.17 | 77.28 | 77.76 | 67.11 | 69.51 | 78.34 | 78.80 | 78.08 | 78.62 | 71.29 | 73.36 |
| | MD | 65.76 | 68.18 | 66.57 | 69.35 | 77.30 | 77.94 | 67.21 | 69.48 | 78.74 | 78.97 | 78.11 | 78.71 | 72.28 | 73.82 |
| | MCTN | 66.19 | 68.58 | 66.70 | 69.01 | 78.32 | 78.41 | 68.10 | 69.34 | 79.11 | 79.14 | 78.65 | 78.64 | 72.85 | 73.94 |
| | MMIN | 67.11 | 68.67 | 67.01 | 69.31 | 78.67 | 78.71 | 68.17 | 69.74 | 79.94 | 79.96 | 79.32 | 79.29 | 73.37 | 74.39 |
| | MPMM | 66.94 | 68.74 | 67.21 | 69.27 | 78.21 | 78.30 | 68.11 | 69.79 | 79.41 | 79.47 | 79.63 | 79.71 | 73.25 | 74.17 |
| | MPLMM | 67.33 | 68.71 | 67.29 | 69.40 | 79.12 | 79.17 | 68.21 | 69.91 | 80.45 | 80.43 | 80.11 | 80.13 | 73.75 | 74.68 |
| | ProMMA | **80.53** | **79.59** | **79.69** | **78.74** | **81.92** | **81.87** | **81.99** | **81.74** | **83.87** | **83.55** | **82.48** | **82.41** | **81.75** | **81.32** |
| CH-SIMS | LB | 63.82 | 75.15 | 64.08 | 78.11 | 76.74 | 76.90 | 62.14 | 73.21 | 76.84 | 76.93 | 77.01 | 77.13 | 70.11 | 76.24 |
| | MS | 62.45 | 74.59 | 63.58 | 76.86 | 77.28 | 77.84 | 60.18 | 71.09 | 76.01 | 76.30 | 77.13 | 77.20 | 69.44 | 75.65 |
| | MD | 64.22 | 77.25 | 63.87 | 76.01 | 77.34 | 77.48 | 62.91 | 72.14 | 76.77 | 76.92 | 77.14 | 77.31 | 70.38 | 76.19 |
| | MCTN | 64.39 | 76.48 | 64.12 | 76.34 | 77.78 | 77.92 | 63.47 | 73.11 | 76.68 | 76.71 | 77.21 | 77.36 | 70.61 | 76.32 |
| | MMIN | 65.21 | 77.09 | 65.32 | 77.41 | 78.91 | 78.67 | 64.28 | 73.36 | 77.32 | 77.33 | 77.40 | 77.48 | 71.41 | 76.89 |
| | MPMM | 64.98 | 76.41 | 65.40 | 77.92 | 78.56 | 78.65 | 64.01 | 73.47 | 77.11 | 77.20 | 77.51 | 77.47 | 71.26 | 76.85 |
| | MPLMM | 65.93 | 77.10 | 66.02 | **78.86** | **79.75** | **78.74** | 65.28 | 74.02 | 77.45 | 77.84 | 77.97 | 77.95 | 72.07 | 77.42 |
| | ProMMA | **73.25** | **77.99** | **72.37** | 78.68 | 73.68 | 78.39 | **73.46** | **74.35** | **78.37** | **78.68** | **78.25** | **78.39** | **74.89** | **77.74** |

TABLE II
ABLATION RESULTS FOR DIFFERENT MODULES

| MME+MiPD | DPW | MPDC | MOSI | | | | MOSEI | | | | CH-SIMS | | | |
|---|---|---|---|---|---|---|---|---|---|---|---|---|---|---|
| | | | ACC | F1 | MAE | CORR | ACC | F1 | MAE | CORR | ACC | F1 | MAE | CORR |
| × | × | × | 71.57 | 71.64 | 108.12 | 60.64 | 78.86 | 79.6 | 56.01 | 74.47 | 77.02 | 77.85 | 46.35 | 54.8 |
| √ | × | × | 71.87 | 71.94 | 108.12 | 56.53 | 79.65 | 79.73 | 60.64 | 68.4 | 77.68 | 79.12 | 48.86 | 48.67 |
| √ | √ | × | 72.71 | 72.88 | 108.04 | 56.62 | 79.67 | 79.79 | 60.52 | 68.5 | 78.12 | 79.61 | 48.88 | 48.56 |
| √ | √ | √ | 72.87 | 73.02 | 107.49 | 56.39 | 81.01 | 80.36 | 61.57 | 67.83 | 78.24 | 79.88 | 47.59 | 52.22 |

*2) Performance under Different Missing Rates:* To assess model robustness under varying degrees of data missingness, we systematically adjusted the missing rate and applied a structured missing strategy consistent with the MPLMM method. The results are presented in Fig.3. Overall, the classification performance of the model demonstrates a "slow decay" pattern across increasing missing rates. In the low missing range (10%–30%), metrics show only a slight decline (generally within 2%), with both accuracy and F1 scores remaining high. This suggests that the MME and MiPD modules effectively identify and mitigate interference from limited missing data, preventing invalid inputs from affecting predictions. Within the medium missing range (40%–60%), a more noticeable decay occurs (approximately 5%), yet the model maintains stable predictions. This robustness can likely be attributed to the multi-level prompting dynamic connection (MPDC) module, which captures global prior knowledge to compensate for missing information. Under high missing rates (70%–90%), the degree of decay becomes dataset-dependent. Performance variation between MOSEI and MOSI indicates that the number of effective samples influences decay severity: fewer samples hinder the learning of stable prediction parameters, as seen in the sharp decline on MOSI. Meanwhile, differences between SIMS and MOSI reveal that feature richness (e.g., visual feature dimensions of 709 in SIMS vs. 20 in MOSI) also governs model stability. Richer features provide more comprehensive complementary information under missing conditions, leading to more consistent predictions. In summary, ProMMA maintains a significant and stable performance advantage in varying missing rates, particularly in extreme missing scenarios.

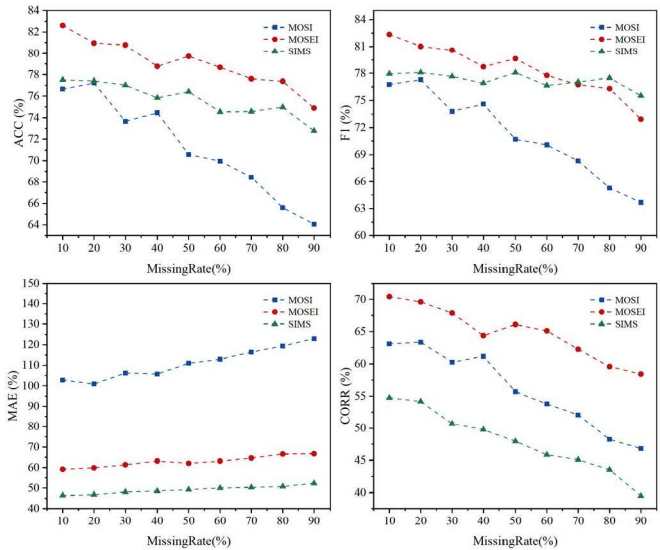

Fig. 3. Model performance variation in ACC, F1, MAE, and CORR under varying missing rates.

## CONCLUSION

This paper introduces a Prompt-based Missing Modality Adaptation framework (ProMMA) to tackle multi-scenario missing modality in multimodal sentiment analysis. The ProMMA framework integrates four key components: a Missing Modality Evaluator (MME), a Modality-invariant Prompt Decoupler (MiPD), a Dynamic Prompt Weighting (DPW) module, and a Multi-level Prompt Dynamic Connection (MPDC) module. Specifically, the MME assesses the importance of a missing modality via a pre-trained model and pseudo-labels to decide whether generation is warranted, thereby avoiding perturbations from low-quality synthetic data. Concurrently, the MiPD module employs prompt decoupling to construct generation prompts, enhancing contextual correlations within modality-specific prompts. To mitigate information loss and interference from missing modalities, the DPW module weights modality feature matrices based on their bimodal mutual information. The MPDC module incorporates global prompt priors through residual connections, refining the granularity of multimodal sentiment representations and thereby enhancing their consistency and completeness. Comparative experiments on benchmark data demonstrate the effectiveness of ProMMA in handling diverse missing-modality scenarios. Furthermore, comprehensive parameter sensitivity analyzes and ablation studies confirm the model's robustness to various missing-modality patterns, as well as the efficacy and synergistic contributions of its constituent modules. Collectively, ProMMA provides a novel theoretical reference for addressing missing-modality challenges in MSA. Specifically, its contributions to prompt decoupling and missing-modality evaluation offer fresh perspectives for future research on interpretable soft prompts and model evaluation.